\documentclass[runningheads]{llncs}
\usepackage{graphicx}
\usepackage{svg}
\usepackage{amsmath}
\usepackage{amsfonts}

\begin{document}
\title{Demystifying Graph Neural Network Explanations}
%
%

\author{Anna Himmelhuber\inst{1,2} \and
Mitchell Joblin\inst{1} \and Martin Ringsquandl\inst{1} \and
Thomas Runkler\inst{1,2} }
\authorrunning{A. Himmelhuber et al.}
%
\institute{Siemens AG, Munich, Germany \\
\email{\{anna.himmelhuber,mitchell.joblin, martin.ringsquandl, thomas.runkler\}@siemens.com}
\and
Technical University of Munich, Munich, Germany}


%
\maketitle              

\begin{abstract}
 Graph neural networks (GNNs) are quickly becoming the standard approach for learning on graph structured data across several domains, but they lack  transparency  in  their  decision-making. Several perturbation-based approaches have been developed to provide insights into the decision making process of GNNs. As this is an early research area, the methods and data used to evaluate the generated explanations lack maturity. We explore these existing approaches and identify common pitfalls in three main areas:  (1) synthetic data generation process, (2) evaluation metrics, and (3) the final presentation of the explanation. For this purpose, we perform an empirical study to explore these pitfalls along with their unintended consequences and propose remedies to mitigate their effects.
\end{abstract}

\section{Introduction}
Many important real-world data sets are graphs or networks. These include social networks, knowledge graphs, protein-protein interaction networks, the World Wide Web and many more. Graph neural networks leverage link structure to encode information as well as incorporate node feature information \cite{zhou2018graph} and currently achieve state-of-the-art on many prediction tasks \cite{xu2018powerful}. Similarly to other connectionist models, GNNs lack transparency in their decision-making. Explaining GNNs is currently in the early stages of research, but since graphs are particularly expressive by encoding contexts, they are a promising candidate when it comes to producing rich explanations \cite{lecue2020role}. The most popular type of GNN explainer methods is perturbation-based, where the output variations are studied with respect to different input perturbations \cite{yuan2020}.\\
\\
When developing any explainable method, it is important to evaluate the performance of the method with respect to valid procedures and metrics \cite{arrieta2020explainable}. With this in mind, we explore the evaluation methods employed by perturbation-based explainer methods in the GNN domain. Validating explanations is generally a challenging task because a ground-truth explanation is not always available. Even for synthetically generated datasets with ground-truth explanations, this approach can be error prone. While validation schemes do exist, they lack maturity. \\
\\
Many explainer methods come with differing evaluation protocols, which makes their comparison difficult. However, some of these protocols overlap or have been adopted by others, e.g. with several papers \cite{cf2021,funke2020,pg2020} using state-of-the-art explainer method \cite{ying2019gnnexplainer} and its evaluation protocol as benchmark. As GNN explainer methods become more and more popular, its vital to avoid evaluation pitfalls and, in the next step, introduce a standard evaluation approach. That's why we perform an empirical study on perturbation-based explanations for GNNs, with  focus on \cite{ying2019gnnexplainer}. Our contributions include identifying pitfalls in three main areas: (1) synthetic data generation process, (2) evaluation metrics, and (3) the final presentation of the explanation. For each pitfall we propose a remedy to increase the validity of the evaluation.

\section{Terminology and Concepts}
\begin{center}
\begin{figure*}
\includegraphics[width=12cm]{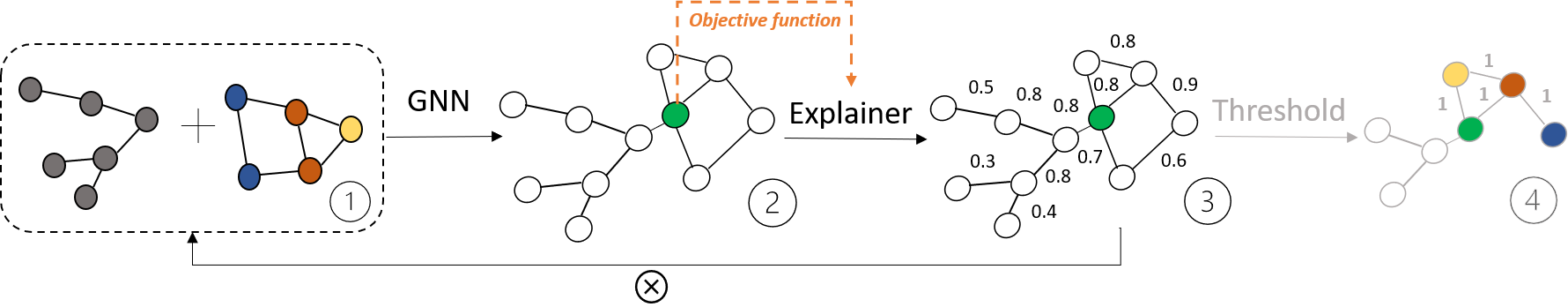}
\caption{Data generation, training and explanation process}
\label{concepts}
\vspace{-2mm}
\end{figure*}
\end{center}

For perturbation-based explainer methods for GNNs, the  output  consists of masks, indicating important input features, including node masks, edge masks or node feature masks depending on the explanation task. We can observe three different types of masks that have been proposed, including soft masks (GNNExplainer \cite{ying2019gnnexplainer}, CF-GNNExplainer\cite{cf2021}),  discrete masks (ZORRO \cite{funke2020}) and approximated discrete masks (PGExplainer \cite{pg2020})\footnote{Please refer to the Appendix~\ref{app3} for more details on GNNs and explainer methods.}. These mask are then applied to the input graph(s) and fed into the trained GNNs to carry out predictions, which is targeted by the objective function to be similar to the original prediction. 
The currently overarching established explanation evaluation scheme for perturbation-based explainer methods consists of a data generation, GNN training, and mask generation step as is shown in Figure \ref{concepts}. The generated synthetic data is comprised of a base graph and a specific motif (1), which are connected randomly and additionally perturbed by noise.  A GNN is applied to the graph execute a prediction task, e.g. node classification (2). In the next step, the explainer method generates masks of the receptive field (3). If the explainer method outputs soft masks, thresholding is needed to arrive at the final explainer subgraph (4). Please refer to the Appendix~\ref{app3} for detailed background on GNNs and  perturbation-based explainer methods.
\subsection{Terminology}\label{app1}
\begin{itemize}
    \item \textbf{Ground-truth explanation}: The ground-truth explanation is a particular motif that is used during the synthetic data set generation, e.g. the "house motif" shown in Figure \ref{concepts} (1). The motif is embedded into a larger graph and perturbed with noise.
    \item \textbf{Ground-truth label}: The ground-truth label is the respective class a node (or graph) is assigned to.
    \item \textbf{Explainer subgraph} with importance scores: The explainer method assigns importance scores to the edges, indicating their influence in the prediction by the GNN, as shown in Figure \ref{concepts} (3).
    \item \textbf{Threshold application}: In order to arrive at a compact subgraph, a threshold is applied to reduce the explainer subgraph to the most important edges by removing all edges that fall below the threshold. 
    \item \textbf{Final (explainer) subgraph}: A reduced final explainer subgraph remains, as shown in Figure \ref{concepts} (4).
    \item \textbf{Label flip}: If the input to the GNN is changed, e.g. using the reduced final explainer subgraph instead of the original receptive field, a label flip can occur. This means that a different class is predicted than in the original prediction. 
\end{itemize}
\subsection{Synthetic Data}
 \textbf{BA-Shapes:}  Node classification dataset with a base graph of 300 nodes and a set of 80 five-node “house”-structured network motifs, which are attached to randomly selected nodes of the base graph and function as ground-truth explanations. The resulting graph is further perturbed by adding 0.1$N$ random edges. Nodes are assigned to 4 classes based on their structural roles.  In a house-structured motif as can be seen in Figure \ref{concepts} (1), there are 3 types of roles: the top (yellow), shoulder (orange) and bottom (blue) node of the house and nodes that do not belong to a house (grey). \\
 \textbf{Tree-Cycles:} Node classification dataset with two different labels, that consists of a base 8-level balanced binary tree and 80 six-node cycle motifs, which are attached to random nodes of the base graph and function as ground-truth explanations.
\section{Pitfalls of Evaluation and Possible Remedies}\label{pitfalls}
\subsection{Pitfall 1: Data Generation}\label{datagen}
\begin{figure*}
\centering
\includegraphics[width=12cm]{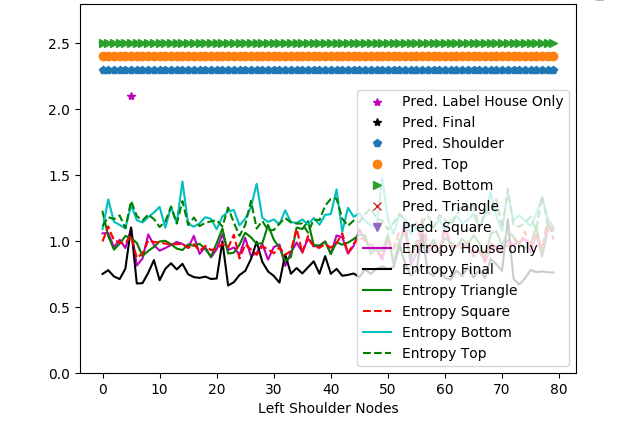}
\caption{Entropies for different motifs and markers (stars) indicating incorrect predictions for the node classification of the left shoulder nodes in the  ``house motif". The markers represent whether the respective motif leads to the correct prediction. A marker is equivalent to the prediction of the incorrect label.}
\label{entropy}
\vspace{-5mm}
\end{figure*}

For all 4 introduced explainer methods \cite{ying2019gnnexplainer,cf2021,funke2020,pg2020}, the synthetic datasets BA-Shapes and Tree-Cycles are used for evaluation. Their advantage is their intuitive motifs and labelling, which is understandable by humans. 
However, the defined ground-truth explanation e.g. the ``house-motif", while being intuitively well-understandable, does not necessarily align with the decision-making process of the GNN and hence doesn't represent the optimal explanation. 
Below, we compare the entropy of the ground-truth explanation to the entropy of other trivial explanation methods including the entire receptive field of the GNN and the target node. Our results\footnote{A 3-layer vanilla Graph Convolutional Network is used carry out experiments.} show the proclaimed ground-truth does not have consistently lower entropy compared to trivial baselines as shown in Table 1.
\begin{table}[h]
\vspace{-2mm}
\label{1}
\begin{center}
\begin{tabular}{ p{4cm}p{2.5cm}p{2.5cm}p{2.5cm} }
 \textbf{Method} &  Top Nodes & Shoulder Nodes &  Bottom Nodes\\
 \hline
Ground-truth  & 1.21    & 0.96     &  0.95   \\
Receptive Field & 1.31 &  0.93 &  1.16 \\   
Target Node   & 1.25    &  1.24 & 1.24\\
\end{tabular}
\end{center}
\caption{Average baseline entropies for BA-Shapes}
\vspace{-5mm}
\end{table}
\\
 For each house node type we see a need for a different type of ground-truth explanation, given the differences in entropy and accuracy performance (see Section \ref{metrics}). Considering Occam's razor, which suggests the simplest explanation is best, we can see for several types of nodes, that the house motif is not the optimal ground-truth explanation. Figure \ref{entropy} (left) shows a number of different possible ground-truth explanations, including the top triangle, the bottom square, the target node ``left shoulder" and right shoulder node, a bottom node, or the top node. It can be seen that while the assigned ground-truth explanation ``house motif" does lead to the correct prediction in nearly all cases, so do the more compact motifs square and triangle with a similarly low entropy and would be the more compact ground-truth explanation.\\
\\
\textbf{Remedy:} We propose a new ground-truth explanation generation in order to better evaluate and compare the explanations. Properties of the dataset should include  the lowest possible entropy matching with the identified ground-truth explanation, e.g. a specific motif. One way to achieve this, is to do a motif search: The entropy and prediction accuracy of several different potential motifs around the target node are calculated, similarly to our approach in Figure \ref{entropy}. The best result is then chosen to be the ground-truth explanation, which ensures maximal compactness and therefore comprehensibility of the explanation. This motif search step implies some additional work and resources, but it is a worthwhile trade-off as it ensures that the ground-truth explanation for a specific class prediction from a specific GNN is known and therefore confirms the validity of the evaluation outcome. It also ensures, that only essential parts of the graph are in the ground-truth explanation.
\subsection{Pitfall 2: Evaluation Metrics}\label{metrics}
, whereas in  \cite{cf2021}, it is the proportion of explanations that are “correct”. In \cite{ying2019gnnexplainer} anFor evaluating the accuracy of an explainer method, the ground-truth explanation has to be known. For synthetic datasets, graph motifs can be used as an approximation ground-truth, even though the GNN might not make predictions as intuitively expected as discussed in Section \ref{datagen}.
When choosing a metric, many papers use the general term  ``accuracy" with wildly differing definitions. In \cite{funke2020}, accuracy is used, defined as the matching
rate for important edges in explanations compared with
those in the ground truthsd \cite{pg2020}, an accuracy is formalized according to a binary classification task, where edges in the ground-truth explanation are treated as labels and the importance scores are viewed as prediction scores. The accuracy is equivalent to the calculation of the area under the curve (AUC) of the receiver operating characteristic curve (ROC), which is calculated on the prediction score and not on the predicted classes. ROC AUC has limitations in its capacity to evaluate the explanations, as it only gives us the probability that a randomly chosen positive instance (edges in the ground-truth explanation) is ranked higher than a randomly chosen negative instance. However, when evaluating an explanation, we do care about the actual probability of the evaluation being correct, instead of just the ranking. Even more problematic is the ROC AUC's tendency to be misleading in situations with high class imbalance \cite{saito2015precision}, as is the case here. Due to the large number of true negatives, edges that are neither in the ground-truth explanation nor in the final explainer subgraph, the false positive rate is pulled down substantially, which leads to an overly optimistic result.\\
\textbf{Remedy:}
Precision-Recall curves (PR) and the corresponding AUC, as opposed to the ROC AUC, can provide a less misleading evaluation due to the fact that they evaluate the fraction of true positives among positive predictions \cite{saito2015precision}. Furthermore, for comparability to hard mask methods and since  for the final explanation presented to the user, a threshold has to be applied, threshold-dependent metrics should be included in the evaluation. We propose to use recall, to account for the sparsity of an explanation. \\
The edges in the ground-truth explanation represent the true positives and the false negatives are the edges that are in the final explainer subgraph but not in the ground-truth explanation as it provides information about the compactness and therefore comprehensibility of the explainer subgraph. Table 2 shows the difference between avg. ROC AUC and avg. PR AUC. As we expect, PR AUC does not achieve the same level as the ROC AUC with a difference of up to 47 percentage points, providing a more comprehensive picture of the explanation quality, similarly to the average recall.
\begin{table}[t]
\label{accu}
\begin{center}
\begin{tabular}{ p{2.5cm}p{1.3cm}p{1.3cm}p{1.5cm}p{1.3cm}p{1.5cm}p{1.3cm} }
 \textbf{Class} &  \textbf{ROC AUC} &  \textbf{SD}& \textbf{PR AUC} {\scriptsize (proposed)}& \textbf{SD} & \textbf{Recall} {\scriptsize (proposed)}& \textbf{SD}  \\
 \hline
Top Nodes  & 0.98 & 0.07 & 0.69 &  0.19 & 0.65 & 0.18 \\
Shoulder Nodes & \textbf{0.98} & 0.91   & \textbf{0.51} & 0.19 & 0.51 & 0.13 \\   
Bottom Nodes   & 0.93 & 0.18 &0.56 & 0.22& 0.57 & 0.21     \\
Cycle Nodes & 0.71 &  0.22  & 0.55 & 0.16  & 0.52 &  0.14 \\
\end{tabular}
\end{center}
\caption{Avg. ROC AUC, Avg. PR AUC and average recall with standard deviations (SD) for GNNExplainer}
\vspace{-5mm}
\end{table}
\subsection{Pitfall 3: Threshold Application}\label{thr}
Reducing the size of the original subgraph is a post-processing step, executed after training the GNN. It is possible that the originally predicted label can flip, resulting in fidelity of the explanation not being ensured. Fidelity refers to an explanation being faithful to the model it aims to explain. 
For example for \cite{ying2019gnnexplainer} a label flip occurs for 66 $\%$ of the top node explainer subgraphs with $T$ = 6. In this case, the final explainer subgraph would lead to a different prediction than the original subgraph and therefore defeating its purpose of explaining the original prediction. Overall, for the BA-Shape dataset, in \cite{ying2019gnnexplainer} 19$\%$ of labels flip, in \cite{cf2021} 39$\%$ of labels flip and in \cite{pg2020} 18$\%$ of labels flip. \\
\\
Additionally, for soft mask explainer methods \cite{ying2019gnnexplainer,cf2021}, the size of the final explainer subgraph is parameterized. In the established evaluation scheme a dedicated hyperparameter $T$ controls the size of the final explainer subgraph. For synthetic datasets, this hyperparameter is set according to knowledge about the ground-truth motif. Using this approach to configure the threshold leads to leaking information and unfairly biases the result. Therefore, the resulting evaluation is flawed, since it can be assumed that the ground-truth size of the explainer subgraph is not typically known. Choosing an appropriate threshold is not trivial, as the resulting recall can differ substantially for different thresholds as can be seen in Table 3, showing a necessary trade-off between the compactness and completeness of the final explainer subgraph. 
\begin{table}[h]
\label{tableacc}
\begin{center}
\begin{tabular}{ p{3cm}p{1.4cm}p{1.4cm}}
 \textbf{Class} &  \textbf{Recall                                  }  th = 6 & \textbf{Recall                                                     }   th = 20 \\
 \hline
Top Nodes  & 0.65 &  0.98 \\
Shoulder Nodes & 0.51  & 0.82  \\   
Bottom Nodes   & 0.57  & 0.75  \\
Cycle Nodes & \textbf{0.52} & \textbf{0.97}\\
\end{tabular}
\end{center}
\caption{Average recall and precision for different thresholds for GNNExplainer}
\vspace{-5mm}
\end{table}
\\
\textbf{Remedy:}  The integration of an additional test to ensure that no label flip occurred for the final explainer subgraph is highly recommended to avoid an explanation that leads to an different decision - in other words, ensuring explanation fidelity. A simple if-then mechanism, that moves on to the next optimal explanation, in case a label flip occurs, would be sufficient.\\
Furthermore, to ensure that no knowledge about the ground truth leaks into the evaluation of the explanation and biases the result for soft mask approaches, we recommend to configure the size of the final explainer subgraph not to a fixed number of edges, but to carry out a grid-search on a test set to choose the optimal threshold.  
\section{Conclusion}
The expressive nature of graphs makes them a promising candidate for producing rich explanations for GNN decision-making. But since a mature standardized approach to evaluating explanations for GNN explainer methods is missing, a valid comparison of different methods can be challenging. For this reason, we find it important to examine existing evaluation methods closely to uncover potential pitfalls. In this paper, we show the implications of three identified evaluation pitfalls in the context of GNNs and propose remedies to avoid them.

\appendix

\section{Background on GNNs and Perturbation-Based Explainer Methods}\label{app3}
For a GNN, the goal is to learn a function of features on a graph \(G=(V, E)\) with edges \(E\) and nodes \(V\). The input is comprised of a feature vector \(x_i\) for every node $i$, summarized in a feature matrix $X$ $\in$ $\mathbb{R}^{n \times d_{in}}$ and a representative description of the link structure in the form of an adjacency matrix $A$. The output of the convolutional layer is a node-level latent representation matrix $Z \in \mathbb{R}^{n \times d_{out}}$, where $d_{out}$ is the number of output latent dimensions per node. Therefore, every convolutional  layer can be written as a non-linear function:
    \[ H^{(l+1)} = f(H^{(l)}, A),\]
    with  \(H^{(0)} = X\) and  \(H^{(L)} = Z\), $L$ being the number of stacked layers. The vanilla GNN model employed here, uses the propagation rule \cite{kipf2016semi}:
    \[ f(H^{(l)}, A) = \sigma(\hat{D}^{-\frac{1}{2}}\hat{A}\hat{D}^{-\frac{1}{2}}H^{(l)}W^{(l)}),\]
    with \(\hat{A} = A + I\), $I$ being the identity matrix. $\hat{D}$ is the diagonal node degree matrix of $\hat{A}$, $W^{(l)}$ is a weight matrix for the $l-th$ neural network layer and $\sigma$  is a non-linear activation function.
Taking the latent node representations $Z$ of the last layer we define the logits of node $v_i$ for a node classification task as follows:    
\[
    \hat{y}_i = \text{softmax}(z_i W^{\top}_{c}),
\]
where $W_c \in \mathbb{R}^{d_{out} \times k}$ projects the node representations into the $k$ dimensional classification space.
\\
\\
\textbf{GNNExplainer:} The GNNExplainer takes a trained GNN and its prediction(s), and it returns an explanation in the form of a small subgraph of the input graph together with a small subset of node features that are most influential for the prediction. For their selection, the mutual information between the GNN prediction and the distribution of possible subgraph structures is maximized through optimizing the conditional entropy. \\
\textbf{CF-GNNExplainer:} The CF-GNNEXPLAINER works by perturbing input data at the instance-level. The instances
are nodes in the graph since it is focused on node classification. The method iteratively
removes edges from the original adjacency matrix based on matrix sparsification techniques, keeping
track of the perturbations that lead to a change in prediction, and returning the perturbation with
the smallest change w.r.t. the number of edges, after adding different edges to the
subgraph.\\
\textbf{ZORRO:} ZORRO employs discrete masks to identify important input nodes and node features through a greedy algorithm, where nodes or node features are selected step by step. The goodness of the explanation is measured by the expected deviation from the prediction of the underlying model.  A subgraph of the
node’s computational graph and its set of features are relevant for a classification decision if the
expected classifier score remains nearly the same when randomizing the remaining features. \\
\textbf{PGExplainer:} The PGExplainer learns approximated discrete masks
for edges to explain the predictions.  Given an input
graph, it first obtains the embeddings for each edge by
concatenating node embeddings. Then the predictor uses
the edge embeddings to predict the probability of each
edge being selected, similarly to an importance
score. The approximated discrete masks are then sampled
via the reparameterization trick. Finally, the objective function maximises the mutual information between
the original predictions and new predictions.

\end{document}